\documentclass{llncs}

\usepackage[utf8]{inputenc}

\usepackage[bookmarks = true, colorlinks=true, linkcolor = black, citecolor = black, menucolor = black, urlcolor = black]{hyperref}
\usepackage{graphicx}
\usepackage{amssymb}
\usepackage{color}
\usepackage{tikz}
\usepackage{verbatim}
\usepackage{theorem}

\newcommand{\id}{Id}
\newcommand{\rrdc}{\mbox{\,\(\Rightarrow\hspace{-9pt}\Rightarrow\)\,}}
\newcommand{\Zset}{\mathbb{Z}}

\theorembodyfont{\rmfamily}
\newtheorem{Thm}{Theorem}

\newtheorem{Defn}[Thm]{Definition}

\newtheorem{Alg}[Thm]{Algorithm}
\newcommand{\inp}{{\itshape Input: }}
\newcommand{\outp}{{\itshape Output: }}
\newcommand{\spec}{{\itshape Description: }}

\newcommand{\sCoq}{{\sc Coq }}
\newcommand{\SSReflect}{{\sc SSReflect }}

\begin{document}

\title{Verifying an algorithm computing Discrete Vector Fields for digital imaging\thanks{Partially supported by Ministerio de Educaci\'on y Ciencia, project MTM2009-13842-C02-01, and by the European Union's
7th Framework Programme under grant agreement nr. 243847 (ForMath).}}

\titlerunning{Verifying an algorithm computing Discrete Vector Fields for digital imaging}  % abbreviated title (for running head)
\author{J\'onathan Heras \and Mar\'ia Poza \and Julio Rubio}
\authorrunning{J. Heras \and M. Poza \and J. Rubio} % abbreviated author list (for running head)
%\tocauthor{author1, author2}
%

\institute{Department of Mathematics and Computer Science of University of La Rioja \\
\email{\{jonathan.heras, maria.poza,  julio.rubio\}@unirioja.es}}

\maketitle              % typeset the title of the contribution

\begin{abstract}

In this paper, we present a formalization of an algorithm to construct \emph{admissible discrete vector fields}
in the \sCoq theorem prover taking advantage of the \SSReflect library. Discrete vector fields are a tool
which has been welcomed in the \emph{homological analysis} of digital images since it provides a procedure
to reduce the amount of information but preserving the homological properties. In particular,
thanks to discrete vector fields, we are able to compute, inside {\sc Coq}, homological properties of
\emph{biomedical images} which otherwise are out of the reach of this system.

\keywords{Discrete Vector Fields, Haskell, {\sc Coq}, {\sc SSReflect}, Integration}

\end{abstract}

\section{Introduction}\label{sec:intro}
Kenzo~\cite{Kenzo} is a Computer Algebra System devoted to Algebraic Topology which was developed by F. Sergeraert. This
system has computed some homology and homotopy groups which cannot be easily obtained by theoretical or computational
means; some examples can be seen in~\cite{RR2012}. Therefore, in this situation, it makes sense to analyze the Kenzo 
programs in order to ensure the correctness of the mathematical results which are obtained thanks to it. To this aim,
two different research lines were launched some years ago to apply formal methods in the study of Kenzo.

On the one hand, the ACL2 theorem prover has been used to verify the correctness of actual Kenzo \emph{programs}, 
see~\cite{LOPSTR10,LMRR2}. ACL2 fits perfectly to this task since Kenzo is implemented in Common Lisp~\cite{Gra96}, the same
language in which ACL2 is built on. Nevertheless, since the ACL2 logic is first-order, the full verification of Kenzo
is not possible, because it uses intensively higher order functional programming. On the other hand, some instrumental
Kenzo \emph{algorithms}, involving higher-order logic, have been formalized in the proof assistants Isabelle/HOL and
Coq. Namely, we can highlight the formalizations of the Basic Perturbation Lemma in Isabelle/HOL, see~\cite{ABR08}, 
and the Effective Homology of Bicomplexes in {\sc Coq}, published in~\cite{DR2011}. 

The work presented in this paper goes in the same direction that the latter approach, formalizing Kenzo \emph{algorithms}.
In particular, we have focused on the formalization of \emph{Discrete Vector Fields}, a powerful notion which will play
a key role in the new version of Kenzo; see the Kenzo web page~\cite{Kenzo}. To carry out this task, we will use the \sCoq
proof assistant~\cite{Coq} and its \SSReflect library~\cite{SSReflect}. 

The importance of Discrete Vector Fields, which were first introduced in~\cite{Forman}, stems from the fact that they
can be used to considerably reduce the amount of information of a discrete object but preserving homological properties.
In particular, we can use discrete vector fields to deal with biomedical images inside \sCoq in a reasonable amount of time.

The rest of this paper is organized as follows. In the next section, we introduce some mathematical preliminaries,
which are encoded abstractly in \sCoq in Section~\ref{sec:foandaics}. Such an abstract version is \emph{refined} to an effective one
in Section~\ref{sec:aeifhtc}; namely, the implementation and formal verification of the main algorithm involved in our 
developments are presented there. In order to ensure the feasibility of our programs, a major issue when applying formal 
methods, we use them to study a biomedical problem in Section~\ref{sec:atbi}. The paper ends with a section of
Conclusions and Further work, and the Bibliography. 

The interested reader can consult the complete development in 
\url{http://wiki.portal.chalmers.se/cse/pmwiki.php/ForMath/ProofExamples#wp3ex5}.

\section{Mathematics to formalize}\label{sec:mtf}
In this section, we briefly provide the minimal mathematical background needed to understand
the rest of the paper. We mainly focus on definitions which, mainly, come from the algebraic setting of 
discrete Morse theory presented in~\cite{cvd} and the Effective Homology theory~\cite{RS02}.
We assume as known the notions of \emph{ring}, \emph{module} over a ring and \emph{module morphism} 
(see, for instance,~\cite{Basic-Algebra}).

First of all, let us introduce one of the main notions in the context of Algebraic Topology: \emph{chain complexes}.

\begin{Defn}
A \emph{chain complex} $C_\ast$ is a pair of sequences $(C_n,d_n)_{n\in\mathbb{Z}}$ where for every $n\in\mathbb{Z}$,
$C_n$ is an ${\mathcal R}$-module and $d_n:C_n\rightarrow C_{n-1}$ is a module morphism, called the \emph{differential map}, 
such that the composition $d_nd_{n+1}$ is null. In many situations the ring ${\mathcal R}$ is either the integer ring,
${\mathcal R} =\mathbb{Z}$, or the field $\mathbb{Z}_2$. In the rest of this section, we will work with $\mathbb{Z}$ as
ground ring; later on, we will change to $\mathbb{Z}_2$. 

\noindent The module $C_n$ is called the module of \emph{$n$-chains}. The image
$B_n = im~d_{n+1} \subseteq C_n$ is the (sub)module of $n$-boundaries. The kernel $Z_n = \ker d_n \subseteq C_n$ is the
(sub)module of \emph{$n$-cycles}.
\end{Defn}

Given a chain complex $C_\ast=(C_n,d_n)_{n \in \mathbb{Z}}$, the identities $d_{n-1}\circ d_n= 0$ mean the inclusion 
relations \(B_n \subseteq Z_n\): every boundary is a cycle (the converse in general is not true). Thus the next definition makes sense.

\begin{Defn}
The \emph{$n$-homology group} of $C_\ast$, denoted by $H_n(C_\ast)$, is defined as the quotient $H_n(C_\ast)=Z_n/B_n$
\end{Defn}

Chain complexes have a corresponding notion of morphism. 

\begin{Defn}
Let $C_\ast=(C_n,d_n)_{n\in\mathbb{Z}}$ and $D_\ast=(D_n,\widehat{d}_n)_{n\in\mathbb{Z}}$ be two chain complexes. A \emph{chain
complex morphism} $f:C_\ast \rightarrow D_\ast$ is a family of module morphisms, $f=\{f_n:C_n \rightarrow D_n\}_{n\in\mathbb{Z}}$,
satisfying for every $n\in\mathbb{Z}$ the relation $f_{n-1}d_n = \widehat{d}_nf_n$. Usually, the sub-indexes are skipped, and we just write
$fd_C = d_Df$.
\end{Defn}

Now, we can introduce one of the fundamental notions in the effective homology theory. 

\begin{Defn}
\label{defn:rdct}
A \emph{reduction} $\rho$ between two chain complexes $C_\ast$ and $D_\ast$, denoted in this paper by $\rho : C_\ast \rrdc D_\ast$, is a triple $\rho=(f,g,h)$
where $f:C_\ast\rightarrow D_\ast$ and $g:D_\ast\rightarrow C_\ast$ are chain complex morphisms, $h = \{h_n:C_n \rightarrow C_{n+1}\}_{n\in\mathbb{Z}}$ 
is a family of module morphism, and the following relations are satisfied:
\begin{enumerate}
\item [1)]$f \circ g = \id_{D_\ast}$;
\item [2)]$d_{C}\circ h + h\circ  d_{C} = \id_{C_\ast}- g\circ f$;
\item [3)]$f \circ h = 0$; \quad $h \circ g = 0$; \quad $h \circ h = 0$.
\end{enumerate}
\end{Defn}

\noindent The importance of reductions lies in the fact that given a reduction $\rho~:~C_\ast \rrdc D_\ast$, then 
$H_n(C_\ast)$ is isomorphic to $H_n(D_\ast)$ for every $n\in \mathbb{Z}$. Very frequently, $D_\ast$ is a much smaller
chain complex than $C_\ast$, so we can compute the homology groups of $C_\ast$ much faster by means of those of $D_\ast$.

Let us state now the main notions coming from the algebraic setting of Discrete Morse Theory~\cite{cvd}. 

\begin{Defn} 
Let $C_\ast=(C_n,d_n)_{n \in \Zset}$ be a free chain complex with distinguished $\Zset$-basis $\beta_n \subset C_n$. 
A \emph {discrete vector field} $V$ on $C_\ast$ is a collection of pairs $V = \{(\sigma_i; \tau_i)\}_{i\in I}$ satisfying the conditions: 
\begin{itemize}
\item Every $\sigma_i$ is some element of $\beta_n$, in which case $\tau_i\in \beta_{n+1}$.
The degree $n$ depends on $i$ and in general is not constant.
\item Every component $\sigma_i$ is a \emph{regular face} of the corresponding $\tau_i$ (regular face means that the coefficient of
$\sigma_i$ in $d_{n+1}\tau_i$ is $1$ or $-1$).
\item Each generator (\emph{cell}) of $C_\ast$ appears at most one time in $V$.
\end{itemize}
\end{Defn}

It is not compulsory all the cells of $C_\ast$ appear in the vector field $V$. 

\begin{Defn}
A cell $\chi$ which does not appear in a discrete vector field $V=\{(\sigma_i; \tau_i)\}_{i\in I}$ is called a \emph{critical cell}.
\end{Defn}

From a discrete vector field on a chain complex, we can introduce $V$-paths. 

\begin{Defn} 
A $V$-path of degree $n$ and length $m$ is a sequence $((\sigma_{i_k}, \tau_{i_k}))_{0 \leq k < m}$ satisfying:
\begin{itemize}
\item Every pair $((\sigma_{i_k}, \tau_{i_k}))$ is a component of $V$ and $\tau_{i_k}$ is a $n$-cell.
\item For every $0 < k < m$, the component $\sigma_{i_k}$ is a face of $\tau_{i_{k-1}}$ (the coefficient of $\sigma_{i_k}$ in $d_{n}\tau_{i_{k-1}}$ is non-null) 
different from $\sigma_{i_{k-1}}$.
\end{itemize}
\end{Defn}

Now we can present the notion of \emph{admissible} discrete vector field on a chain complex, a concept which can be understood as 
a \emph{recipe} indicating both the ``useless'' elements of the chain complex (in the sense, that they can be removed without 
changing its homology) and the \emph{critical} ones (those whose removal modifies the homology).

\begin{Defn} 
A discrete vector field $V$ is \emph{admissible} if for every $n \in \Zset$, a function $\lambda_n: \beta_n \rightarrow \mathbb{N}$ is
provided satisfying the following property: every $V$-path starting from $\sigma \in \beta_n$ has a length bounded by $\lambda_n(\sigma)$.
\end{Defn}

Finally, we can state the theorem where Discrete Morse Theory and Effective Homology converge. 

\begin{Thm}~\cite[Theorem~19]{cvd}\label{thm:advred}
Let $C_\ast=(C_n,d_n)_{n \in \Zset}$ be a free chain complex and $V$ be an admissible discrete vector
field on $C_\ast$. Then the vector field $V$ defines a canonical reduction $\rho : (C_n, d_n)\rrdc (C^c_n, d^c_n)$ where $C^c_n = \Zset[\beta^c_n]$
is the free $\Zset$-module generated by $\beta^c_n$, the critical $n$-cells.
\end{Thm} 

\noindent Therefore, as the bigger the admissible discrete vector field $V$ the smaller the chain complex $C^c_\ast$, we need algorithms
which produce admissible discrete vector fields as large as possible. 

If we consider the case of \emph{finite type} chain complexes, where there is a finite number of generators in each dimension of the chain complex,
the differential maps can be represented as matrices. In that case, the problem of finding an admissible discrete vector field on the chain complex 
can be solved through the computation of an admissible vector field for those matrices. 

\begin{Defn}\label{def:dvfM}
Let $M$ be a matrix with coefficients in $\Zset$, and with $m$ rows and $n$ columns. A discrete vector field $V$ 
for this matrix is a set of natural pairs $\{(a_i, b_i)\}$ satisfying these conditions:
\begin{enumerate}
 \item $1\leq a_i \leq m$ and $1\leq b_i \leq n$.
 \item The entry $M[a_i,b_i]$ of the matrix is $\pm 1$.
 \item The indexes $a_i$ (resp. $b_i$) are pairwise different.
\end{enumerate}
\end{Defn}

Given $V$ be a vector field for our matrix $M$, we need to know if $V$ is admissible. If $1\leq a,a'\leq m$, with $a\neq a'$, we can
decide $a > a'$ if there is an elementary $V$-path from $a$ to $a'$, that is, if a vector $(a,b)$ is present in $V$ and the entry $M[a',b]$
is non-null. In this way, a binary relation is obtained. Then the vector field $V$ is admissible if and only if this binary relation 
generates a partial order, that is, if there is no loop $a_1 > a_2 > \ldots > a_k = a_1$.

Eventually, given a matrix and an admissible discrete vector field on it, we can construct a new matrix, smaller than the original one,
preserving the homological properties. This is the equivalent version of Theorem~\ref{thm:advred} for matrices, a detailed description
of the process can be seen in~\cite[Proposition~14]{cvd}. 

In the rest of the paper, we will focus on the formally certified construction of an admissible discrete vector field from a matrix.
The task of verifying the reduction process remains as further work.

\section{A non deterministic algorithm in \SSReflect}\label{sec:foandaics}
First of all, we have provided in {\sc Coq}/\SSReflect an abstract formalization of admissible discrete vector fields on matrices
and a non deterministic algorithm to construct an admissible discrete vector field from a matrix\footnote{Thanks
are due to Maxime D\'en\`es and Anders M\"ortberg which guided us in this development.}. {\sc SSReflect}
is an extension for the {\sc Coq} proof assistant, which was developed by G. Gonthier while formalizing 
the Four Color Theorem~\cite{FCT}. Nowadays, it is used in the formal proof of the Feit-Thompson theorem~\cite{MathComp}.

{\sc SSReflect} provides all the necessary tools to achieve our goal. In particular, we take
advantage of the \verb"matrix", \verb"ssralg" and \verb"fingraph" libraries, which formalize,
respectively, matrix theory, the main algebraic structures and the theory of finite graphs. 
 
First of all, we are going to define admissible discrete vector field on a matrix $M$ with coefficients in
a ring $\mathcal{R}$, and with $m$ rows and $n$ columns. It is worth noting that our matrices
are defined over a generic ring instead of working with coefficients in $\mathbb{Z}$ since
the {\sc SSReflect} implementation of $\mathbb{Z}$, see~\cite{CM11}, is not yet included in the \SSReflect distributed
version. The vector fields are represented by a sequence of pairs where the first component is an ordinal $m$ and the second one an
ordinal $n$.  

\begin{verbatim}
Variable R : ringType.
Variables m n : nat.
Definition vectorfield := seq ('I_m * 'I_n).
\end{verbatim}

Now, we can define in a straightforward manner a function, called \verb"dvf", which given a matrix \verb"M" (with coefficients in
a ring $\mathcal{R}$, and with $m$ rows and $n$ columns, \verb"'M[R]_(m,n)") and an object \verb"V" of type \verb"vectorfield"
checks whether \verb"V" satisfies the properties of a discrete vector field on \verb"M" (Definition~\ref{def:dvfM}). 

\begin{verbatim}
Definition dvf (M : 'M[R]_(m,n)) (V : vectorfield) :=
  all [pred p | (M p.1 p.2 == 1) || (M p.1 p.2 == -1)] V && 
  (uniq (map (@fst _ _) V) && uniq (map (@snd _ _) V)).
\end{verbatim}

\noindent It is worth noting that the first condition of Definition~\ref{def:dvfM} is implicit in the \verb"vectorfield" type. Now, as we have
explained at the end of the previous section, from a discrete vector field \verb"V" a binary relation is obtained between the first elements of each pair
of \verb"V". Such a binary relation will be encoded by means of an object of the following type.

\begin{verbatim}
Definition orders := (simpl_rel 'I_m).
\end{verbatim}

Finally, we can define a function, which is called \verb"advf", that given a matrix \verb"'M[R]_(m,n)", \verb"M", a \verb"vectorfield",
\verb"V" and an \verb"orders", \verb"ords", as input, tests whether both \verb"V" satisfies the properties of a discrete vector field 
on \verb"M" and the admissibility property for the relations, \verb"ords", associated with the vector field, \verb"V". In order to test the admissibility
property we generate the transitive closure of \verb"ords", using the \verb"connect" operator of the \verb"fingraph" library,
and subsequently check that there is not any path between the first element of a pair of \verb"V" and itself. 

\begin{verbatim}
Definition advf (M:'M[R]_(m,n)) (V:vectorfield) (ords:orders) :=
 dvf M V && all [pred i|~~(connect ords i i)] (map (@fst _ _) V). 
\end{verbatim}

Now, let us define a non deterministic algorithm which construct an admissible discrete vector field from a matrix. Firstly,
we define a function, \verb"gen_orders", which generates the relations between the elements of the discrete vector
field as we have explained at the end of the previous section. 

\begin{verbatim}
Definition gen_orders (M0 : 'M[R]_(m,n)) (i:'I_m) j :=
  [rel i x | (x != i) && (M0 x j != 0)].
\end{verbatim}

Subsequently, the function, \verb"gen_adm_dvf", which generates an admissible discrete vector field from a matrix is introduced.
This function invokes a recursive function, \verb"genDvfOrders", which in each step adds a new component to the vector field
in such a way that the admissibility property is fulfilled. The recursive algorithm stops when either there is not any new element 
whose inclusion in the vector field preserves the admissibility property or the maximum number of elements of the discrete vector field
(which is the minimum between the number of columns and the number of rows of the matrix) is reached.  

\begin{verbatim}
Fixpoint genDvfOrders M V (ords : simpl_rel _) k :=
  if k is l.+1 then 
    let P := [pred ij | admissible (ij::V) M 
                        (relU ords (gen_orders M ij.1 ij.2))] in
    if pick P is Some (i,j) 
       then genDvfOrders M ((i,j)::V)
                         (relU ords (gen_orders M i j)) l
    else (V, ords)
  else (V, ords).

Definition gen_adm_dvf M := 
  genDvfOrders M [::] [rel x y | false] (minn m n).
\end{verbatim}

Eventually, we can certify in a straightforward manner (just 4 lines) the correctness of the function \verb"gen_adm_dvf". 

\begin{verbatim}
Lemma admissible_gen_adm_dvf m n (M : 'M[R]_(m,n)) :
 let (vf,ords) := gen_adm_dvf M in admissible vf M ords.
\end{verbatim}

As a final remark, it is worth noting that the function \verb"gen_adm_dvf" is not executable. On the one
hand, \SSReflect matrices are locked in a way that do not allow direct computations since they may trigger
heavy computations during deduction steps. On the other hand, we are using the \verb"pick" instruction, in 
the definition of \verb"genDvfOrders", to choose the elements which are added to the vector field; however,
this operator does not provide an actual method to select those elements.

\section{An effective implementation: from Haskell to \sCoq}\label{sec:aeifhtc}

In the previous section, we have presented a non deterministic algorithm to construct an admissible discrete vector field from a matrix. 
Such an abstract version has been described on high-level datastructures; now, we are going to obtain from it a \emph{refined} version,  
based on datastructures closer to machine representation, which will be executable. 

The necessity of an executable algorithm which construct an admissible discrete vector field stems from the fact that they will
will play a key role to study biomedical images. There are several algorithms to construct an admissible discrete vector field; 
the one that we will use is explained in~\cite{cvd} (from now on, called RS's algorithm; RS stands for Romero-Sergeraert). The 
implementation of this algorithm will be executable but the proof of its correctness will be much more difficult than the one 
presented in the previous section.

\subsection{The Romero-Sergeraert algorithm}\label{subsec:trsa}

\begin{comment}
{This algorithm returns an admissible discrete vector field built in the following way. It consists of running the rows of the matrix $M$ in the usual reading order looking for the entries whose value is $1$. This entry will be in a position of $M$, for instance $(i,j)$ which will be exactly the vector to add to the admissible discrete vector field. Let us note that every the entries can not be selected because of the properties of an admissible discrete vector field depicted in Section \ref{sec:mtf}. Moreover, to avoid that cycles are not created, that is to say, satisfy the admissibility property, we have to pay attention to the entries of the column \emph{j} different from $0$ which are in a different position from \emph{i}.}
\end{comment}

The underlying idea of the RS algorithm is that given an admissible discrete vector field, we try to enlarge it adding new vectors which 
preserve the admissibility property. We can define algorithmically the RS algorithm as follows.

\begin{Alg}[The RS Algorithm]\label{alg:rs}
\textcolor{white}{.}\\
\inp a matrix $M$ with coefficients in $\mathbb{Z}$.\\
\outp an admissible discrete vector field for $M$.\\
\spec
  \begin{enumerate}
    \item Initialize the vector field, $V$, to the void vector field.
    \item Initialize the relations, $ords$, to nil.
    \item For every row, $i$, of $M$:
      \begin{enumerate}
        \item Search the first entry of the row equal to $1$ or $-1$, $j$.
        \item \verb"If" $(i,j)$ can be added to the vector field; that is, if we add it to $V$ and generate all
              the relations, the properties of an admissible discrete vector field are preserved. 
        \begin{itemize}
          \item[] \verb"then": 
                     \begin{itemize}
                        \item[-] Add $(i,j)$ to $V$.
                        \item[-] Add to $ords$ the corresponding relations generated from $(i,j)$.
                        \item[-] Go to the next row and repeat from Step 3.
                      \end{itemize}
          \item[] \verb"else": look for the next entry of the row whose value is $1$ or $-1$.
                     \begin{itemize}
                        \item[-] \verb"If" there is not any.
                         \begin{itemize}
                           \item[] \verb"then": go to the next row and repeat from Step 3.
                           \item[] \verb"else": go to Step 3.2 with $j$ the column of the entry whose value is $1$ or $-1$.
			\end{itemize}
                      \end{itemize}
        \end{itemize}
      \end{enumerate}
  \end{enumerate}
\end{Alg}

In order to clarify how this algorithm works, let us construct an admissible discrete vector field from the following matrix. 

 \[ \left( \begin{array}{cccc}
1 & 1 & 0 & 0  \\
1 & 1 & 1 & 0 \\
0 & 0 & 1 & 1 \\
0 & 1 & 1 & 0
 \end{array} \right)\]

We start with the void vector field $V= \left\{\right\}$. Running the successive rows, we find $M\left[1,1\right] = 1$, and we include the
vector $(1,1)$ to $V$, obtaining $V = \left\{(1,1)\right\}$. Then, let us add the relations that, in this case is $1 > 2$ because 
$M\left[2,1\right]\neq 0$. So, it will be forbidden to incorporate the relation $2>1$ as the cycle $1 > 2 > 1$ would appear. 
Besides, the row 1 and the column 1 are now used and cannot be used anymore. So, we go on with the second row and find $M\left[2,1\right] = 1$,
but we cannot add $(2,1)$ as we have just said. Moreover, the element $(2,2)$ can not be incorporated because the cycle $1 > 2 > 1$ would be created.
So, we continue and find the next element, $M\left[2,3\right] = 1$. This does not create any cycle and satisfies the properties of a discrete vector field.
Then, we obtain $V=\left\{(2,3),(1,1)\right\}$ and the relations $2 > 3$ and $2 > 4$. Running the next row, the first element equal to $1$ is in the position $(3,3)$, 
but we cannot include it due to the admissibility property. Therefore, we try with the last element of this row $(3,4)$. No relation is generated in this case
because in this column the only non null element is in the chosen position. So, $V=\left\{(3,4),(2,3),(1,1)\right\}$. Finally, we run the last row. 
The elements that could be added are $(4,2),(4,3)$, but in both cases we would have to append the relation $4>2$. This would generate a cycle with one 
of the previous restrictions, $2>4$. So, we obtain $V= \left\{(3,4),(2,3),(1,1)\right\}$ and the relations are: $1>2$, $2>3$ and $2>4$.

In general, Algorithm~\ref{alg:rs} can be applied over matrices with coefficients in a general ring ${\cal R}$. From now on, we will work with
${\cal R}=\mathbb{Z}_2$, since this is the usual ring when working with monochromatic images in the context of Digital Algebraic Topology. 

The development of a formally certified implementation of the RS algorithm has followed the methodology presented in~\cite{and10}. Firstly,
we implement a version of our programs in \emph{Haskell}~\cite{haskell98}, a \emph{lazy} functional programming language. Subsequently, we
intensively test our implementation using \emph{QuickCheck}~\cite{Quickcheck}, a tool which allows one to automatically test properties
about programs implemented in Haskell. Finally, we verify the correctness of our programs using
the \sCoq \emph{interactive} proof assistant and its \SSReflect library.

\subsection{A Haskell program}

The choice of Haskell to implement our programs was because both the code and the programming style is similar to the
ones of \sCoq. In this programming language, we have defined the programs which implement the RS algorithm. The
description of the main function is shown here:

\begin{description}
\item[{\small gen\_admdvf\_ord $M$:}]
From a matrix $M$ with coefficients in $\mathbb{Z}_2$, represented as a list of lists, this function generates
an admissible discrete vector field for $M$, encoded by a list of natural pairs, and the relations, a list of lists of
natural numbers.
\end{description}

Let us emphasize that the function \verb"gen_admdvf_ord" returns a pair of elements. The former one, \verb"(gen_admdvf_ord M).1",
is a discrete vector field and the latter one, \verb"(gen_admdvf_ord M).2", corresponds to the relations associated with the vector field.
To provide a better understanding of these tools, let us apply them in the example presented in Subsection~\ref{subsec:trsa}.

\begin{verbatim}
> gen_admdvf_ord [[1,1,0,0],[1,1,1,0],[0,0,1,1],[0,1,1,0]]
[([(3,4),(2,3),(1,1)], [[2,4],[2,3],[1,2],[1,2,4],[1,2,3]])]
\end{verbatim}

Let us note that we return the transitive closure of the relations between the first components of the pairs of the discrete vector field. This will
make the proof of the correctness of our programs easier. 
 
\subsection{Testing with QuickCheck}\label{subsec:twq}

Using QuickCheck can be considered as a good starting point towards the formal verification of our programs.
On the one hand, a specification of the properties which must be satisfied by our programs is given (a necessary
step in the formalization process). On the other hand, before trying a formal verification of our programs
(a quite difficult task) we are testing them, a process which can be useful in order to detect bugs.

In our case, we want to check that the output by \verb"gen_admdvf_ord" gives us an admissible discrete vector field.
Then, let $M$ be a matrix over $\mathbb{Z}_2$ with $m$ rows and $n$ columns, $V=(a_i, b_i)_i$ be a discrete 
vector field from $M$ and $ords$ be the transitive closure of the relations associated with $V$, the properties to test are
the ones coming from Definition~\ref{def:dvfM} and the admissibility property adapted to the $\mathbb{Z}_2$ case.

\begin{enumerate}
  \item $1\leq a_i \leq m$ and $1\leq b_i \leq n$.
  \item $\forall i\;, M(a_i,b_i) = 1$.
  \item $(a_i)_i$ (resp. $(b_i)_i$) are pairwise different.
  \item $ords$ does not have any loop (admissibility property).
\end{enumerate}

These four properties has been encoded in Haskell by means of a function called \verb"isAdmVecfield". To test in QuickCheck
that our implementation of the RS algorithm fulfills the specification given in \verb"isAdmVecfield", the following
\emph{property definition}, using QuickCheck terminology, is defined. 

\begin{verbatim}
condAdmVecfield M = 
let advf = (gen_admdvf_ord M) in isAdmVecfield M (advf.1) (advf.2)
\end{verbatim}

The definition of \verb"condAdmVecfield" states that given a matrix \verb"M", the output
of \verb"gen_admdvf_ord", both the discrete vector field (first component) and the relations (second component) from $M$, fulfill 
the specification of the property called \verb"isAdmVecfield". Now, we can test whether \verb"condAdmVecfield" satisfies such 
a property.

\begin{verbatim}
> quickCheck condAdmVecfield
+ + + OK, passed 100 tests.
\end{verbatim}

The result produced by QuickCheck when evaluating this statement, means that QuickCheck has generated $100$ random values for \verb"M", 
checking that the property was true for all these cases.

\subsection{Formalization in \sCoq/\SSReflect}\label{subsec:fics}

After testing our programs, and as a final step to confirm their correctness, we can undertake the challenge of formally verify them.

First of all, we define the data types related to our programs, which are effective matrices, vector fields and relations. 
We have tried to keep them as close as possible to the Haskell ones; therefore, a matrix is represented by means of a list  
of lists over $\mathbb{Z}_2$, a vector field is a sequence of natural pairs and finally, the relations is a 
list of lists of natural numbers. 

\begin{verbatim}
Definition Z2 := Fp_fieldType 2.
Definition matZ2 := seq (seq Z2).
Definition vectorfield := seq (prod nat nat).
Definition orders := seq (seq nat).
\end{verbatim}

Afterwards, we translate both the programs and the properties, which were specified during the testing of the programs,
from Haskell to {\sc Coq}, a task which is quite direct since these two systems are close.

\begin{comment}Let us emphasize that any modification in the code {\sc Coq} regarding to Haskell version has to be modified in this one. Moreover, the algorithms will have to be tested again to check that these changes do not affect to the hopeful result.\end{comment}

Then, we have defined a function \verb"isAdmVecfield" which receives as input a matrix over $\mathbb{Z}_2$, a vector field and the relations and
checks if the properties, explained in Subsection~\ref{subsec:twq}, are satisfied. 
\begin{verbatim}
Definition isAdmVecfield (M:matZ2)(vf:vectorfield)(ord:orders):=
  ((longmn (size M) (getfirstElemseq vf)) /\
  (longmn (size (nth [::] M 0)) (getsndElemseq vf))) /\
  (forall i j:nat, (i , j) \in vf -> compij i j M = 1) /\
  ((uniq (getfirstElemseq vf)) /\ (uniq (getsndElemseq vf))) /\
  (admissible ord).
\end{verbatim}

Finally, we have proved the theorem \verb"genDvfisVecfield" which says that given a matrix $M$, the output produced by 
\verb"gen_admdvf_ord" satisfies the properties specified in \verb"isAdmVecfield".

\begin{verbatim}
Theorem genDvfisVecfield (M:matZ2): 
  let advf := (gen_admdvf_ord M) in 
  isAdmVecfield M (advf.1) (advf.2).
\end{verbatim}

\noindent We have split the proof of the above theorem into $4$ lemmas which correspond with each one of the properties that should
be fulfilled to have an admissible discrete vector field. For instance, the lemma associated with the first property of the definition of a 
discrete vector field is the following one.

\begin{verbatim}
Lemma propSizef (M:matZ2):
  let advf := (gen_admdvf_ord M).1 in
  (longmn (size M) (getfirstElemseq advf) /\
  (longmn (size (nth nil M 0))(getsndElemseq advf)).
\end{verbatim}

Both the functions which implement the RS algorithm and the ones which specify the definitional properties of admissible discrete vector fields
are defined using a \emph{functional style}; that is, our programs are defined using \emph{pattern-matching} and \emph{recursion}. Therefore,
in order to reason about our recursive functions, we need elimination principles which are fitted for them. To this aim, 
we use the tool presented in~\cite{BC02} which allows one to reason about complex recursive definitions since \sCoq does not directly generate
elimination principles for complex recursive functions. Let us see how the tool presented in~\cite{BC02} works.

In our development of the implementation of the RS algorithm, we have defined a function, called \verb"subm", which takes as arguments
a natural number, \verb"n", and a matrix, \verb"M", and removes the first \verb"n" rows of \verb"M". The inductive
scheme associated with \verb"subm" is set as follows. 

\begin{verbatim}
Functional Scheme subm_ind := Induction for subm Sort Prop.
\end{verbatim}

Then, when we need to reason about \verb"subm", we can apply this scheme with the corresponding parameters using the instruction
\verb"functional induction". However, as we have previously said both our programs to define the RS algorithm and the ones which
specify the properties to prove are recursive. Then, in several cases, it is necessary to merge several inductive schemes to induction
simultaneously on several variables.  For instance, let $M$ be a matrix and $M'$ be a submatrix of $M$ where we have removed the 
$(k-1)$ first rows of $M$; then, we want to prove that $\forall j,\; M (i, j) = M' (i - k + 1, j)$. This can be stated in \sCoq as follows.

\begin{verbatim}
Lemma Mij_subM (i k: nat) (M: matZ2):
  k <= i -> k != 0 -> let M' := (subm k M) in
  M i j ==  M' (i - k + 1) j.
\end{verbatim}

To prove this lemma it is necessary to induct simultaneously on the parameters \verb"i", \verb"k" and \verb"M", but 
the inductive scheme generated from \verb"subm" only applies induction on \verb"k" and \verb"M". Therefore, 
we have to define a new recursive function, called \verb"Mij_subM_rec", to provide a proper inductive scheme to prove this theorem. 

\begin{verbatim}
Fixpoint Mij_subM_rec (i k: nat) (M: matZ2) :=
match k with
|0 => M
|S p => match M with
        |nil => nil
        |hM::tM => if (k == 1)
                    then a::b
                  else (Mij_subM_rec p (i- 1) tM)
        end
end.
\end{verbatim}
 
This style of proving functional programs in \sCoq is the one followed in the development of the proof of Theorem \verb"genDvfisVecfield".

\subsection{Experimental results}

Using the same methodology presented throughout this section, we are working in the formalization of the algorithm which, from a matrix and an
admissible discrete vector field on it, produces a reduced matrix preserving the homological properties of the original one. Up to now,
we have achieved a Haskell implementation which has been both tested with QuickCheck and translated into {\sc Coq}; however, the proof of
its correctness remains as further work. 

Anyway, as we have a \sCoq implementation of that procedure, we can execute some examples inside this proof assistant. Namely, 
we have integrated the programs presented in this paper with the ones devoted to the computation of homology groups of digital images
introduced in~\cite{HDMMPS}; and we have considered matrices coming from $500$ randomly generated images. 

The size of the matrices associated with those images was initially around $100\times 300$, and after the reduction process the average
size was $5\times 50$. Using the original matrices \sCoq takes around 12 seconds to compute the homology from the matrices; on the contrary,
using the reduced matrices \sCoq only needs milliseconds. Furthermore, as we will see in the following section, we have studied some images
which are originated from a real biomedical problem.

\section{Application to biomedical images}\label{sec:atbi}
Biomedical images are a suitable benchmark for testing our programs. On the one hand, the amount of information
included in this kind of images is usually quite big; then, a process able to reduce those images but keeping
the homological properties can be really useful. On the other hand, software systems dealing with biomedical
images must be trustworthy; this is our case since we have formally verified the correctness of our programs. 

As an example, we can consider the problem of counting the number of \emph{synapses} in a neuron. Synapses~\cite{Neuroscience}
are the points of connection between neurons and are related to the computational capabilities of the brain. Therefore,
the treatment of neurological diseases, such as Alzheimer, may take advantage of procedures modifying the number
of synapses~\cite{C11}. 

\begin{figure}
\centering
 \includegraphics[scale=0.14]{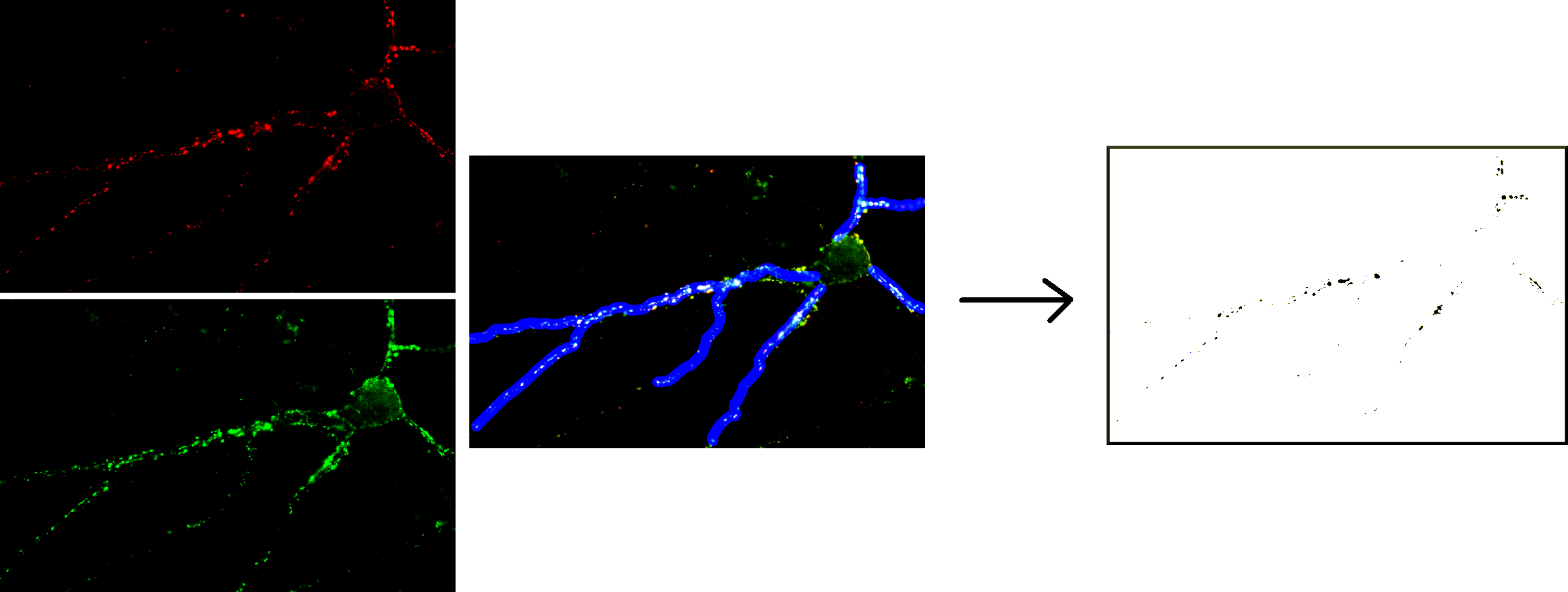}
\caption{Synapses extraction from three images of a neuron}\label{fig:neuron}
\end{figure}

Up to now, the study of the \emph{synaptic density evolution} of neurons was a time-consuming task since it was performed, mainly,
manually. To overcome this issue, an automatic method was presented in~\cite{HMPR11-ACA}. Briefly speaking, such a process can be
split into two parts. Firstly, from three images of a neuron (the neuron with two antibody markers and the structure of the neuron),
a monochromatic image is obtained, see Figure~\ref{fig:neuron}\footnote{The same images with higher resolution can be seen in \url{http://www.unirioja.es/cu/joheras/synapses/}}.
In such an image, each connected component represents a synapse. So, the problem of measuring the number of synapses
is translated into a question of counting the connected components of a monochromatic image.

In the context of Algebraic Digital Topology, this issue can be tackled by means of the computation of the homology group $H_0$ 
of the monochromatic image. This task can be performed in \sCoq through the formally verified programs 
which were presented in~\cite{HDMMPS}. Nevertheless, such programs are not able to handle images like the one of the right side 
of Figure~\ref{fig:neuron} due to its size (let us remark that \sCoq is a proof assistant tool and not a computer algebra system). In order
to overcome this drawback, as we have explained at the end of the previous section, we have integrate our reduction programs with
the ones presented in~\cite{HDMMPS}. Using this approach, we can successfully compute the homology of the biomedical images in just $25$ seconds,
a remarkable time for an execution inside {\sc Coq}.

\section{Conclusions and Further work}
In this paper, we have given the first step towards the formal verification of a procedure 
which allows one to study homological properties of big digital images inside {\sc Coq}. The underlying idea consists
in building an admissible discrete vector field on the matrices associated with an image and,
subsequently reduce those matrices but preserving the homology. 

Up to now, we have certified the former step of this procedure, the construction of an admissible discrete
vector field from a matrix, both in an abstract and a concrete way. The reason because the abstract formalization is
useful is twofold: on the one hand, it provides a high-level theory close to usual mathematics, and, on the other hand,
it has been refined to obtain the effective construction of admissible discrete vector fields. As we have explained, there are several
heuristics to construct an admissible discrete vector field from a matrix, the one that we have
chosen is the RS algorithm~\cite{cvd} which produces, as we have experimentally seen, quite large discrete
vector fields, a desirable property for these objects. The latter step, the process to reduce the matrices,
is already specified in {\sc Coq}, but the proof of its correctness is still an ongoing work. 

The suitability of our approach has been tested with several examples coming from randomly generated images
and also real images associated with a biomedical problem, the study of synaptic density evolution. The results
which have been obtained are remarkable since the amount of time necessary to compute homology groups 
of such images inside \sCoq is considerably reduced (in fact, it was impossible in the case of biomedical images).

As further work, we have to deal with some formalization issues. Namely, we have to verify that a reduction
can be constructed from a matrix and an admissible discrete vector field to a reduced matrix. Moreover, 
we have hitherto worked with matrices over the ring $\mathbb{Z}_2$; the more general case
of matrices with coefficients in a ring ${\cal R}$ (with convenient constructive properties) 
should be studied.

As we have seen in Subsection~\ref{subsec:fics}, it is necessary the definition of inductive schema
which fits to our complex recursive programs. Then, this opens the door to an integration between
\sCoq and the ACL2 Theorem Prover~\cite{ACL2}. ACL2 has good heuristics to generate inductive schemes
from recursive functions; so, we could translate our functional programs from \sCoq to ACL2, generate
the inductive schemes in ACL2; and finally return such inductive schemes to {\sc Coq}. Some preliminary
experiments have been performed to automate that process, obtaining encouraging results.  

In a different research line, we can consider the study of more complex biomedical problems
using our certified programs. As an example, the recognition of the structure of neurons seems to
involve the computation of homology groups in higher dimensions; a question which could be tackled
with our tools.

\bibliographystyle{abbrv}
\bibliography{vaacdvffdi}

\end{document}